\documentclass[10pt, journal, twoside]{IEEEtran}
\ifCLASSINFOpdf
\else
\fi
\usepackage{graphicx}
\usepackage{amsmath}
\usepackage{amssymb}
\usepackage{amsfonts}
\usepackage{booktabs}
\usepackage{multirow}
\usepackage{subcaption} 

\usepackage{eucal}
\usepackage{mathabx}
\usepackage{stmaryrd}
\usepackage{color,xcolor}
\usepackage{cite}
\usepackage{mathrsfs}
\usepackage{fancyhdr}
\usepackage{setspace}
\usepackage{color}

\usepackage{geometry}
\begin{document}
%
\title{Multi-Stage Spatio-Temporal Aggregation Transformer for Video Person Re-identification
}
%
%

\author{Ziyi Tang,
        Ruimao Zhang,~\IEEEmembership{Member,~IEEE,}
        Zhanglin Peng,
        Jinrui Chen,  
        Liang Lin,~\IEEEmembership{Senior Member,~IEEE}
\thanks{Ziyi Tang, Ruimao Zhang, and Jinrui Chen are with The Chinese University of Hong Kong (Shenzhen), and Ziyi Tang is also with Sun Yat-sen University (e-mail: tangziyi@cuhk.edu.cn, ruimao.zhang@ieee.org, and 120090765@link.cuhk.edu.cn ).} 
\thanks{Zhanglin Peng is with the Department of Computer Science, The University of Hong Kong, Hong Kong, China (e-mail: zhanglin.peng@connect.hku.hk ).}
%
%

\thanks{Liang Lin is with the School of Computer Science and Engineering, Sun Yat-sen University (e-mail: linliang@ieee.org).}
\thanks{This paper was done when Ziyi Tang was working as a Research Assistant at The Chinese University of Hong Kong (Shenzhen).}
\thanks{The Corresponding Author is Ruimao Zhang}

}

%
%



\pagestyle{empty}
\markboth {Submission to IEEE Transaction on multimedia}{}



\maketitle
\pagestyle{fancy} 
      \fancyhead[LO]{\scriptsize \thepage} 
      \fancyhead[RO]{\scriptsize \MakeUppercase{IEEE TRANSACTIONS ON MULTIMEDIA}}
      
      \fancyhead[LE]{\scriptsize \MakeUppercase{TANG \MakeLowercase{\textit{et al.}}: Multi-Stage Spatio-Temporal Aggregation Transformer for Video Person Re-identification}} 
      \fancyhead[RE]{\scriptsize \thepage} 
      \cfoot{}
      
      \renewcommand{\headrulewidth}{0pt} 
      \renewcommand{\footrulewidth}{0pt} 


\begin{abstract}
In recent years, the Transformer architecture has shown its superiority in the video-based person re-identification task.
Inspired by video representation learning, these methods mainly focus on designing modules to extract informative spatial and temporal features.
However, they are still limited in extracting local attributes and global identity information, which are critical for the person re-identification task. 
In this paper, we propose a novel Multi-Stage Spatial-Temporal Aggregation Transformer (MSTAT) with two novel designed proxy embedding modules to address the above issue. 
Specifically, MSTAT consists of three stages to encode the attribute-associated, the identity-associated, and the attribute-identity-associated information from the video clips, respectively, achieving the holistic perception of the input person. 
We combine the outputs of all the stages for the final identification.
In practice, to save the computational cost, the Spatial-Temporal Aggregation (STA) modules are first adopted in each stage to conduct the self-attention operations along the spatial and temporal dimensions separately.
We further introduce the Attribute-Aware and Identity-Aware Proxy embedding modules (AAP and IAP) to extract the informative and discriminative feature representations at different stages.
All of them are realized by employing newly designed self-attention operations with specific meanings.
Moreover, temporal patch shuffling is also introduced to further improve the robustness of the model.
Extensive experimental results demonstrate the effectiveness of the proposed modules in extracting the informative and discriminative information from the videos, 
and illustrate the MSTAT can achieve state-of-the-art accuracies on various standard benchmarks.
%
%
%
%
%
%
%
%

\end{abstract}


%
\IEEEpeerreviewmaketitle

\section{Introduction}

\begin{figure}[t]
	\begin{center}
		\includegraphics[width=\linewidth]{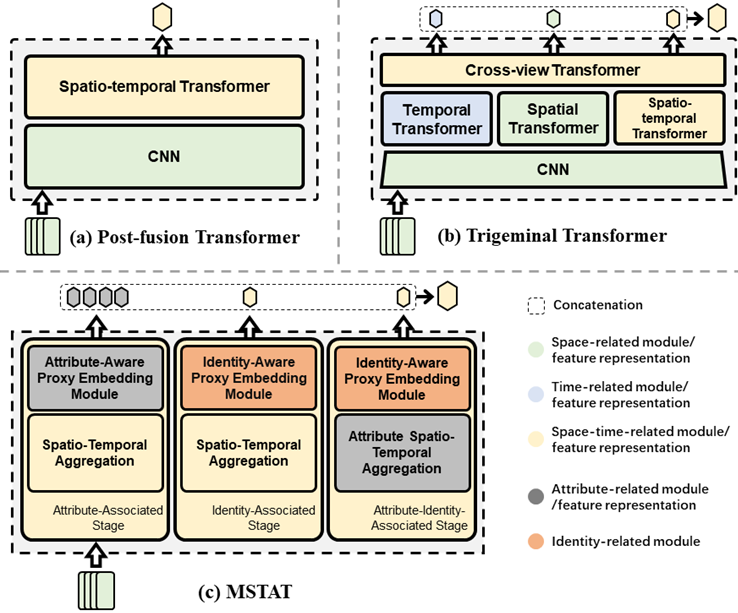}
	\end{center}
	\vspace{-10pt}
	\caption{Comparison between different Transformer-based frameworks for video re-ID. (a) shows the framework where the Transformer fuse post-CNN features of the entire video. (b) is Trigeminal Transformer~\cite{liu2021video}, including three separate streams for temporal,  spatial, and spatio-temporal feature extraction. (c) displays a multi-stage spatio-temporal aggregation Transformer, which consists of three stages, all with a spatio-temporal view but different meanings. 
	}
	\label{fig:method_comparison}
\end{figure}


%
\IEEEPARstart{P}{erson}  Re-identification (re-ID)~\cite{C:bak2017one,C:farenzena2010person,C:gray2008viewpoint}, which aims at matching pedestrians across different camera views at different times, is a critical task of visual surveillance. 
In the earlier stage, the studies have mainly focused on image-based person re-ID~\cite{C:gray2008viewpoint, C:farenzena2010person, C:liu2012person}, which mine the discriminative information in the spatial domain. 
With the development of the monitoring sensors,  multi-modality information has been introduced to re-ID task~\cite{wu2021person, hao2021cross, wu2017robust}. Numerous methods have been proposed to break down barriers between modalities regarding their image styles~\cite{ye2020visible}, structural features~\cite{ye2020dynamic, zhang2022modality, ye2021dynamic}, or network parameters~\cite{ye2020cross, hao2021cross}.
%
%
%

On the other hand, some studies have exploited multi-frame data and proposed various schemes~\cite{C:Viewpoint_Invariant,C:subramaniam2019,C:zhao2017} to extract informative temporal representations to pursue video-based person re-ID.
%
%
%
In such a setting, each time a non-labeled query tracklet clip is given, its discriminative feature representation needs to be extracted to retrieve the clips of the corresponding person in the non-labeled gallery. 
In practice, how to simultaneously extract such discriminative information from spatial and temporal dimensions is the key to improving the accuracy of video-based re-ID.

To address such an issue, traditional methods~\cite{cheng2016person} usually employ hierarchically convolutional architectures to update local patterns progressively. 
Furthermore, some attempts~\cite{zhang2019scan,liu2019spatially,J:chen2020learning,J:chen2019spatial,wu2018and} adopt attention-based modules to dynamically infer discriminative information from videos. 
For instance, Wu \MakeLowercase{\textit{et al.}}~\cite{wu2021person} embed body part prior knowledge inside the network architecture via dense and non-local region-based attention.
%
%
Although recent years have witnessed the success of convolution-based methods~\cite{C:li2014deepreid,C:wu2016enhanced,cheng2016person,hou2020temporal,zhou2019omni,C:chen2018video,zhang2019scan,chen2022saliency}, they have encountered a bottleneck of accuracy improvement, as convolution layers suffer from their intrinsic limitations of spatial-temporal dependency modeling and information aggregation~\cite{zhang2021vidtr}. 


%
Recently, the Transformer architecture~\cite{C:dosovitskiy2020image,liu2021swin,C:TNT,C:T2T} has attracted much attention in the computer vision area because of its excellent context modeling ability. 
The core idea of such a model is to construct interrelationships between local contents via global attention operation. 
%
%
In the literature, some hybrid network architectures~\cite{he2021transreid, chen2021oh, liu2021video} have been proposed to tackle long-range context modeling in video-based re-ID.
A widely used paradigm is to leverage Transformer as the post-processing unit, coupled with a convolutional neural network (CNN) as the basic feature extractor.
For example, as summarized in Fig.~\ref{fig:method_comparison} (a), He \textit{et al.}~\cite{he2021dense} and Zhang \textit{et al.}~\cite{zhang2021spatiotemporal} adopt a monolithic Transformer to fuse frame-level CNN feature. 
As shown in Fig.~\ref{fig:method_comparison} (b), 
Liu \MakeLowercase{\textit{et al.}}~\cite{liu2021video} take a step further and put forward multi-stream Transformer architecture in which each stream emphasizes a particular dimension of the video features.
%
%
%
%
In a hybrid architecture, however, the 2D CNN bottom encoder restricts the long-range spatio-temporal interactions among local contents, which hinders the discovery of contextual cues. 
%
%
Later, to address this problem, some pure Transformer-based approaches are introduced to video-based re-ID. 
Nevertheless, the existing Transformer-based frameworks are mainly motivated by those in video understanding and concentrate on designing the architecture to learn spatial-temporal representations efficiently. 
Most of them are still limited in extracting informative and human-relevant discriminative information from the video clips, which are critical for large-scale matching tasks~\cite{zhou2019omni,zhang2019densely,zhang2020ordered,hou2019vrstc}.

To address the above issues, we propose a novel Multi-stage Spatial-Temporal Aggregation Transformer framework, named MSTAT, which consists of \textbf{three stages} to respectively encode the attribute-associated, the identity-associated, and the attribute-identity-associated information from video clips. 
Firstly, to save the computational cost, the \textbf{Spatial-Temporal Aggregation} (STA) modules~\cite{arnab2021vivit,bertasius2021space}
are firstly adopted in each stage as their building blocks to conduct the self-attention operations along the spatial and temporal dimensions separately. 
Further, as shown in Fig.~\ref{fig:method_comparison}, we introduce the plug-and-play \textbf{Attribute-Aware Proxy} and \textbf{Identity-Aware Proxy}  (AAP and IAP) embedding modules into different stages, for the purpose of reserving informative attribute features and aggregating discriminative identity features respectively. 
They are both implemented by self-attention operations but with different learnable proxy embedding schemes. 
For the AAP embedding module, AAPs play the role of attribute queries to reserve a diversity of implicit attributes of a person. 
Arguably, the combination of these attribute representations is informative and provides discriminative power, complementary to the identity-only prediction. 
%
%
In contrast, the IAP embedding module maintains a group of IAPs as key-value pairs. With explicit constraints, they learn to successively match and aggregate the discriminative identity-aware features embedded in patch tokens. 
During similarity measurement, the output feature representations of the three stages are concatenated to form a holistic view of the input person. 







In practice, a Transformer-specific data augmentation scheme, Temporal Patch Shuffling, is also introduced, which randomly rearranges the patches temporally. With such a scheme, the enriched training data effectively improve the ability to learn invariant appearance features, leading to the robustness of the model. 
%
%
Extensive experiments on three public benchmarks demonstrate our proposed framework is superior to the state-of-the-art on different metrics. Concretely, we achieve the best performance of $91.8\%$ rank-1 accuracy on MARS, which is the largest video re-ID dataset at present. 

In summary, our contributions are three-fold. (1) We introduce a Multi-stage Spatial-Temporal Aggregation Transformer framework (MSTAT) for video-based person re-ID. Compared to existing Transformer-based frameworks, MSTAT better learns informative attribute features and discriminative identity features.
(2) For different stages, we devise two different proxy embedding modules, named Attribute-Aware and Identity-Aware Proxy embedding modules, to extract informative attribute features and aggregate discriminative identity features from the entire video, respectively. 
(3) A simple yet effective data augmentation scheme, referred to as Temporal Patch Shuffling, is proposed to consolidate the network's invariance to appearance shifts and enrich training data.

\section{Related Works}
\label{sec:Related}

\subsection{Image-Based Person Re-ID}
Image-based person re-ID mainly focuses on person representation learning. 
Early works focus primarily on carefully designed handcraft features~\cite{C:bak2017one,C:farenzena2010person,C:gray2008viewpoint,C:liu2012person, C:liao2015person,C:zheng2015towards}. Recently, The flourishing deep learning has become the mainstream method for learning representation in person ReID~\cite{C:li2014deepreid, C:wu2016enhanced, C:xiao2016learning, C:varior2016gated,yu2021apparel,wan2019concentrated}. For the last few years, CNN has been a widely-used feature extractor~\cite{C:li2019multi, C:li2014deepreid, C:liao2015person, C:liao2018video, C:varior2016gated, xia2019second, zhang2019scan, C:yan2016, chen2022knowledge}. 
OSNet~\cite{zhou2019omni} fuses multi-scale features in an attention-style sub-network to obtain informative omni-scale features. 
Some works ~\cite{zhang2019densely, yu2020devil, chen2021cross} focus on extracting and aligning semantic information to address misalignment caused by pose/viewpoint variations, imperfect person detection, etc. 
To avoid the misleading by noisy labels, Ye \textit{et al.}\cite{ye2021collaborative} presents a self-label refining strategy, deeply integrating annotation optimization and network training. 
%
%
So far, some works~\cite{he2021transreid, chen2021oh} also explore Image-based person re-ID based on Vision Transformer~\cite{C:dosovitskiy2020image}. 
For example, TransReID~\cite{he2021transreid} adopts Transformer as the backbone and extracts discriminative features from randomly sampled patch groups. 
%

%

\begin{figure*}[t]
	\begin{center}
		\includegraphics[width=0.9\linewidth]{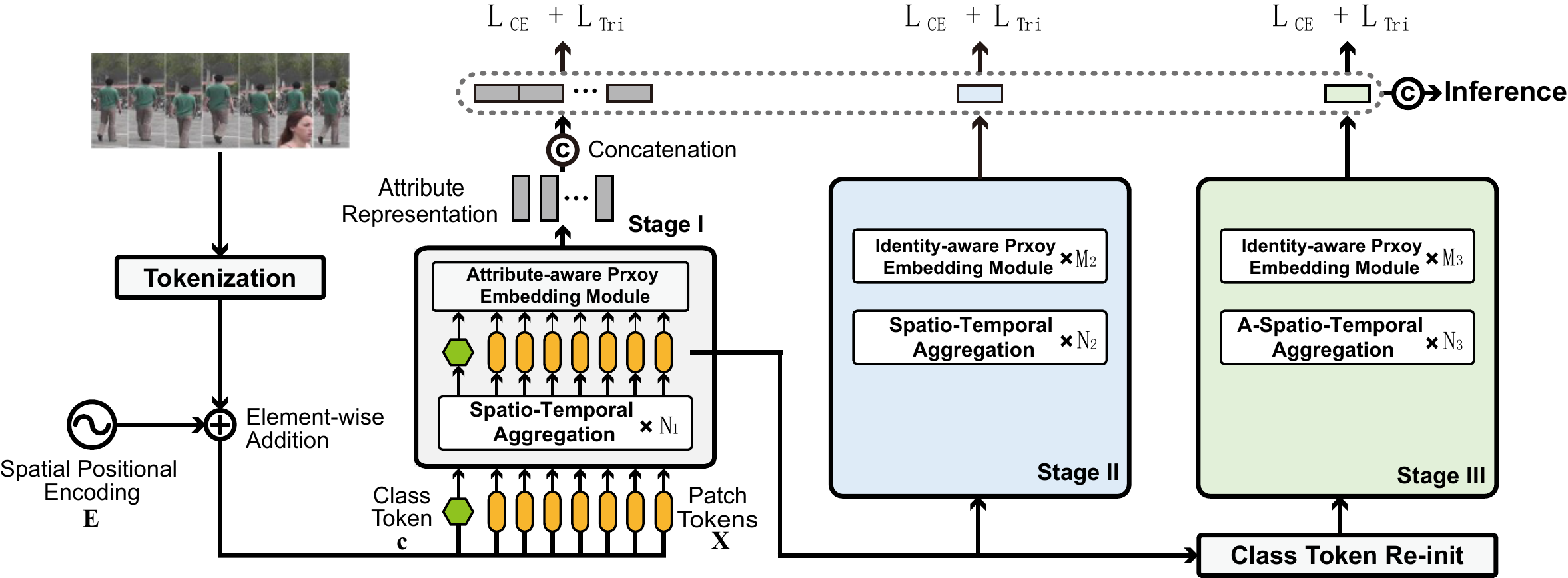}
	\end{center}
	\caption{
	The overall architecture of our proposed MSTAT which consists of three stages, all based on the Transformer architecture. 
	\textbf{Stage I} updates the spatio-temporal patch token sequence of the input video and aggregates them into a group of attribute-associated representations. 
    Subsequently, \textbf{Stage II} aggregates discriminative identity-associated features and \textbf{Stage III} attribute-identity-associated features, relying upon their stage-specific class tokens. 
	Here, we omit the input and output of each module except the attribute-aware proxy embedding module in \textbf{Stage I}.
	%
    %
    At inference time, all these feature representations are combined through concatenation to infer the pedestrian's identity jointly.}
	\label{fig:frmework}
\end{figure*}

\subsection{Video-Based Person ReID}
Compared to image-based person re-ID, video-based person re-ID usually performs better because it provides temporal information and mitigates occlusion by taking advantage of multi-frame information. For capturing more robust and discriminative representation from frame sequences, traditional video-based re-ID methods usually focus on two areas: 1) encoding of temporal information; 2) aggregation of temporal information. 

To encode additional temporal information, early methods~\cite{C:Viewpoint_Invariant,C:subramaniam2019,C:zhao2017} directly use temporal information as additional features. Some works~\cite{C:mclaughlin2016,C:yan2016,J:liu2017video,wu2018and} use recurrent models, \textit{e.g.}, RNNs~\cite{mikolov2010recurrent} and LSTM~\cite{hochreiter1997long}, to process the temporal information. Some other works~\cite{C:chen2018video, C:liu2019, C:zhou2017, C:yan2016, C:mclaughlin2016,chen2022saliency,shi2020person} go further by introducing the attention mechanism to apply dynamic temporal feature fusion. Another class of works~\cite{C:chung2017two} introduces optical flow that captures temporal motion. What is more, some works~\cite{C:zhang2018multi, C:li2018diversity, C:suh2018part, C:wu2018exploit, C:li2018unsupervised, C:zheng2016mars} directly implement spatio-temporal pooling to video sequences and generate a global representation via CNNs. Recently, 3D CNNs ~\cite{C:liao2018video,C:gu2020appearance} learn to encode video features in a joint spatio-temporal manner. 
M3D\cite{C:li2019multi} endows 2D CNN with multi-scale temporal feature extraction ability via multi-scale 3D convolutional kernels.

For the sake of aggregation that aims to generate discriminative features from full video features, a class of approaches~\cite{C:mclaughlin2016,C:zhou2017,zhang2020learning} applies average pooling on the time dimension to aggregate spatio-temporal feature maps. 
Recently, some attention-based methods~\cite{J:chen2020learning, C:li2018diversity, C:xu2017jointly, wu2021person} attained significant performance improvement by dynamically highlighting different video frames/regions so as to filter more discriminative features from these critical frames/regions. For instance, Liu \MakeLowercase{\textit{et al.}}~\cite{liu2021video} introduce cross-attention to aggregate multi-view video features by pair-wise interaction between these views.
Apart from the exploration of more effective architectural design, a branch of works study the effect of pedestrian attributes~\cite{zhao2019attribute, song2019two, chai2022video}, such as \textit{shoes, bag, and down color}, or the gait~\cite{chang2021seq, nambiar2019gait}, \textit{i.e.} walking style of pedestrians, as a more comprehensive form of pedestrian description. 
Chang \MakeLowercase{\textit{et al.}}~\cite{chang2021seq} closely integrate two coherent tasks: gait recognition and video-based re-ID by using a hybrid framework including a set-based gait recognition branch. 
Some works~\cite{zhao2019attribute, song2019two} embed attribute predictors into the network supported by annotations obtained from a network pretrained on an attribute dataset. 
Chai \textit{et al.}~\cite{chai2022video} separate attributes into ID-relevant and ID-irrelevant ones and propose a novel pose-invariant and motion-invariant triplet loss to mine the hardest samples considering the distance of pose and motion states.  
%


%
Although the above methods have made significant progress in performance,  Transformer~\cite{vaswani2017attention}, which is deemed a more powerful architecture to process sequence data, may raise the performance ceiling of video-based re-ID. 
%
To illustrate this, Transformer can readily adapt to video data with the support of the global attention mechanism to capture spatio-temporal dependencies and temporal positional encoding to order spatio-temporal positions.
In addition, the class token is off-the-shelf for Transformer-based models to aggregate spatio-temporal information.
%
However, Transformer suffers from multiple drawbacks~\cite{C:dosovitskiy2020image, C:T2T, C:wang2021pyramid, A:zhang2020feature}, and few works have been released so far on video-based person re-ID based on Transformer. In this work, we attempt to explore the potential of intractable Transformer in video-based person re-ID. 

\subsection{Vision Transformer}
Recently, Transformer has shown its ability as an alternative to CNN. Inspired by the great success of Transformer in natural language processing, recent researchers~\cite{C:dosovitskiy2020image, liu2021swin, C:wang2021pyramid, liu2021swin} have extended Transformer to CV tasks and obtained promising results. 
%

%

Bertasius $et\ al.$~\cite{bertasius2021space} explores different video self-attention schemes considering their cost-performance trade-off, resulting in a conclusion that the divided space-time self-attention is optimal. Similarly, ViViT~\cite{arnab2021vivit} factorizes self-attention to compute self-attention spatially and then temporally. Inspired by these works, we divide video self-attention into spatial attention followed by temporal attention, and we further propose a attribute-aware variant for video-based re-ID. Furthermore, little research has been done on Transformer for Video-based person re-ID. Trigeminal Transformers (TMT)~\cite{liu2021video} puts the input patch token sequence through a spatial, a temporal, and a spatio-temporal minor Transformer, respectively, and a cross-view interaction module fuses their outputs. Differently, MSTAT has three stages, all extracting spatio-temporal features but with different meanings: (1) attribute features, (2) identity features, (3) attribute-identity features.


\section{Method}\label{sec:Method}
In Sec.~\ref{sec:overview}, we first overview the proposed MSTAT framework. Then, Spatio-Temporal Aggregation (STA), the normal spatial-temporal feature extractor in MSTAT, is formulated in section Sec.~\ref{sec:sta}. 
Along with it, we introduce the proposed Attribute-Aware Proxy (AAP) and Identity-Aware Proxy (IAP) embedding modules in Sec.~\ref{sec:aap}. Finally, Temporal Patch Shuffling (TPS), a newly introduced Transformer-specific data augmentation scheme, is presented in section ~\ref{sec:ts}. 


\subsection{Overview}\label{sec:overview}
This section briefly summarizes the workflow of MSTAT. The overall MSTAT framework is shown in Fig.~\ref{fig:frmework}. 
Given a video tracklet $\textbf{V} \in \mathbb{R} ^ {T \times 3 \times H \times W} $ with $T$ frames and the resolution of each frame is $H \times W$, the goal of MSTAT is to learn a mapping from a video tracklet $\textbf{V}$ to a $d$-dimension representation space in which each identity is discriminative from the others.

Specifically, as shown on the left of Fig.~\ref{fig:frmework}, MSTAT first linearly projects non-overlapping image patches of size $3 \times P \times P$ into $d$-dimensional patch tokens, where $d = 3P^2$ denotes the embedded dimension of tokens. 
Thus, a patch token sequence $\textbf{X} \in \mathbb{R} ^ {T \times N \times d}$ is obtained, where the number of patch tokens in each frame is denoted by $ N = \frac{H \times W}{P^2}$. Meanwhile, spatial positional encoding $ \textbf{E} \in \mathbb{R} ^ {N \times d}$ is added to $\textbf{X}$ in a element-wise manner for reserving spatial structure in each frame. 
Notably, we do not insert temporal positional encoding into $\textbf{X}$, since the temporal order is usually not conducive to video-based re-ID, which is also demonstrated in~\cite{zhang2020ordered}.
Finally, a class token $\textbf{c} \in  \mathbb{R} ^ {d}$ is associated with $\textbf{X}$ to aggregate global identity representation. 
%

Next, we feed the token sequence $\textbf{X}$ into \textbf{Stage I} of MSTAT.
It takes $\textbf{X}$ and $\textbf{c}$ as input, and employs a stack of eight Spatio-Temporal Aggregation (STA) blocks for inter-frame and intra-frame correlation modeling.
The output tokens are then fed into an Attribute-Aware Proxy (AAP) embedding module to mine rich visual attributes, a composite group of semantic cues that imply identity information, \textit{e.g.}, {\ttfamily garments}, {\ttfamily handbags} and so on. 
The \textbf{Stage II} includes a series of STA blocks (three in our experiments), followed by an Identity-Aware Proxy (IAP) embedding module which is able to screen out discriminative identity-associated information by inspecting the entire sequence in parallel. 
In the \textbf{Stage III}, we first introduce a novel class token to directly aggregate higher-level features. 
In addition, a stack of Attribute-STA (A-STA) blocks is used to fuse attributes from different frames. 
At last, an IAP embedding module is adopted to generate a discriminative representation for the person.
In the training phase, the attribute representations extracted from \textbf{Stage I} and the class tokens of \textbf{Stage II} and \textbf{Stage III} are supervised separately by a group of losses. 
During the testing, the attribute representations and the class tokens from the last two stages are concatenated for similarity measurement.

\subsection{Spatio-temporal Aggregation}\label{sec:sta}
To begin with, we make a quick review of the vanilla Transformer self-attention mechanism first proposed in ~\cite{vaswani2017attention}. 
In practice, visual Transformer embeds an image into a sequence of patch tokens, and self-attention operation first linearly projects these tokens to the corresponding query $\textbf{Q}$, key $\textbf{K}$ and value $\textbf{V}$ respectively. 
%
%
Then, the scaled product of $\textbf{Q}$ and $\textbf{K}$ generates an attention map $\textbf{A}$, indicating estimated relationships between token representations in $\textbf{Q}$ and $\textbf{K}$. 
Then, $\textbf{V}$ performs a re-weighting by multiplying the attention map $\textbf{A}$, to obtain the output of Transformer self-Attention. 
In this way, patch tokens are reconstructed by leveraging interaction with each other. Formally, self-Attention operation ${\rm SA(.)}$ can be formulated as follows:
%
%
%
\begin{gather}
\textbf{Q}, \textbf{K}, \textbf{V} = \hat{\textbf{S}}\textbf{W}_q, \hat{\textbf{S}}\textbf{W}_k, \hat{\textbf{S}}\textbf{W}_v \notag \\
\textbf{A} = {\rm Softmax}(\textbf{QK}^{\rm T})/\sqrt{d}             \notag \\
{\rm SA}(\hat{\textbf{S}}) = \textbf{AV} \label{eq:1}
\end{gather}
\noindent where $\hat{\textbf{S}} \in \mathbb{R} ^ {\hat{N} \times d}$ denotes an 2-dimensional input token sequence, and $W_q$, $W_k$, $W_v \in \mathbb{R} ^ {d \times d^{\prime}} $ denote three learnable parameter matrices of size $d \times d^{\prime}$.
In the multi-head setting, we let $d^{\prime} = d/n$, where $n$ indicates the number of attention heads. 
The function ${\rm Softmax(\cdot)}$ denotes the softmax operation for each row. 
And the scaling operation in Eqn.(\ref{eq:1}) eliminates the influence from the scale of embedded dimension $d^{\prime}$.  

In our \textbf{Spatio-Temporal Aggregation block} (STA),
self-attention operation along time axis and along space axis (\textit{i.e.} temporal attention and spatial attention) are separately denoted as ${\rm SA_t}(\cdot)$ and ${\rm SA_s}(\cdot)$. 
Let $\textbf{S} \in \mathbb{R} ^ {\hat{T} \times \hat{N} \times d}$ denote an input spatio-temporal token sequence. 
Formally, ${\rm SA_t}(\cdot)$ and ${\rm SA_s}(\cdot)$ can be written as:
\begin{gather}
\small
{\rm SA_t}(\textbf{S}) = {\rm SA}( \rm Concat( \textbf{S}_{:, 0},..., \textbf{S}_{:, n},..., \textbf{S}_{:, N-1} ) ) \notag \\
\small
{\rm SA_s}(\textbf{S}) = {\rm SA}( \rm Concat( \textbf{S}_{0, :},..., \textbf{S}_{t, :},..., \textbf{S}_{T-1, :})) \label{eq:2}
\end{gather} 
where $T$ indicates the total number of frames in video clip, $N$ is the total spatial position index, and ${\rm Concat}(\cdot)$ denotes the concatenation operation in the split dimension, \textit{e.g.}, the spatial position dimension in Eqn.(\ref{eq:2}).

\begin{figure}[t]
	\begin{center}
		\includegraphics[width=0.8\linewidth]{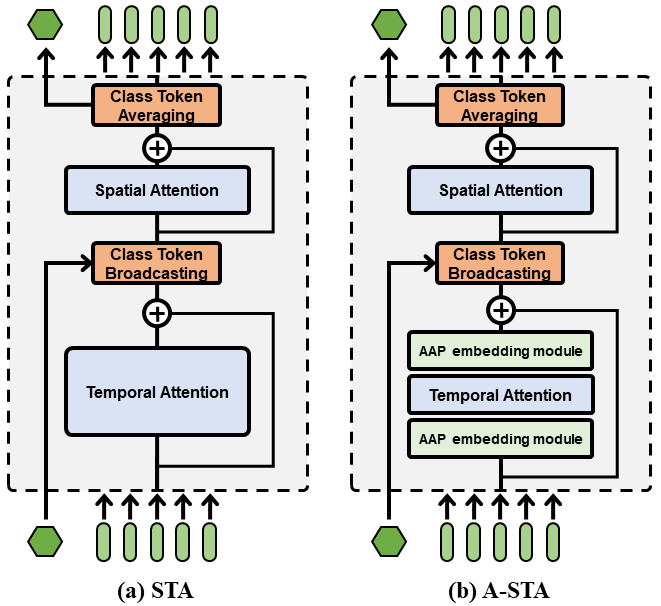}
	\end{center}
	\caption{The detailed comparison between (a) Spatio-Temporal Aggregation block (STA) and (b) Attribution Spatio-Temporal Aggregation block (A-STA). Two additional Attribute-Aware Proxy (AAP) embedding modules are placed into the latter, before and after the temporal attention module.
    The class token broadcasting operation duplicates the class token for each frame to attend spatial attention within a specific frame. 
    Oppositely, class token averaging calculates the average of all class token copies. 
    Note that the Pre-Norm~\cite{xiong2020layer} layers before temporal attention and spatial attention are omitted. }
	\label{fig:sta}
\end{figure}
%
%
%
%
%
%
%
Given ${\rm SA_t}(.)$ and ${\rm SA_s}(.)$, the STA block consecutively integrates these two self-attention modules to extract spatial-temporal features. 
As illustrated in Fig.~\ref{fig:sta}, STA further extracts discriminative information from patch tokens to the class token through spatial attention ${\rm SA_s}(\cdot)$, 
which can be realized by concatenating the copies of class token to the token sequence of each frame before ${\rm SA_s}(.)$, and taking the average of class token copies after ${\rm SA_s}(.)$ to further apply the later temporal aggregation. 
In this way, the general form of STA can be presented as: 
%
\begin{gather}
\small
\textbf{S}^{\prime} = \textbf{S} + \alpha \times {\rm SA_t}({{\rm LN}}(\textbf{S})) \notag \\
{\rm STA}(\textbf{S}, \textbf{c}) = {\rm Concat}(\textbf{S}^{\prime}, \textbf{c})  \notag \\ 
+ \beta \times {\rm SA_s}({\rm LN}({\rm Concat}(\textbf{S}^{\prime}, \textbf{c}) )) 
\label{eq:STA}
\end{gather}

where $\rm LN(\cdot)$ denotes Layer Normalization~\cite{ba2016layer}. 
The hyper-parameter $\alpha$ and $\beta$ are learnable scalar residual weights to balance temporal attention and spatial attention. 
%
%
%
%
%
%
%
%
%
Compared with the space-time joint attention in ~\cite{bertasius2021space} and \cite{arnab2021vivit}, which jointly processes all patches of a video, 
STA is more computation-efficient by reducing complexity from ${\mathbb{O} }(T^2N^2)$ to ${\mathbb O}(T^2+N^2)$. 
Actually, it avoids operating on a long sequence, whose length always leads to quadratic growth of computational complexity~\cite{2020Linformer,guo2021beyond}. 

\begin{figure}[t]
	\begin{center}
		\includegraphics[width=0.6\linewidth]{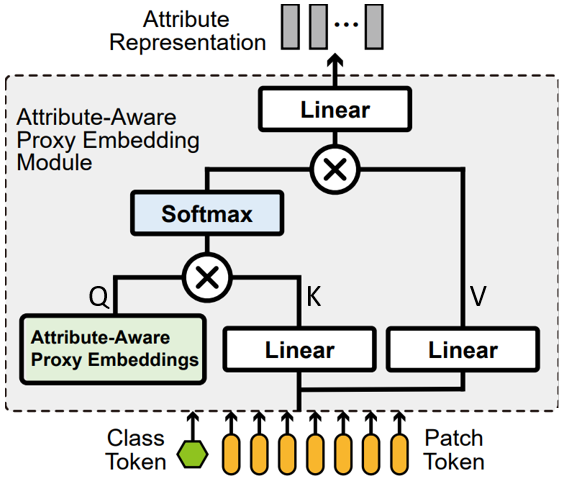}
	\end{center}
	\caption{The detailed module design of the Attribute-Aware Proxy (AAP) embedding module. The Attribute-Aware Proxy Embedding denotes a learnable matrix that is used as the query of the attention operation.
	For simplicity, this figure only shows the single-head version of the AAP embedding module and the scaling operation before the softmax operation is omitted. } 
	\label{fig:attribute-aware}
\end{figure}

\subsection{Attribute-Aware Proxy Embedding Module}
\label{sec:aap}
Local patch tokens usually contain rich attributive information, \textit{e.g.,} {\ttfamily glasses}, {\ttfamily umbrellas}, {\ttfamily logos}, and so on.
Even if a single attribute is not discriminative enough to recover one's identity, the combinations of a pedestrian's rich attributes should be discriminative as each attribute eliminates a certain degree of uncertainty. 
Rather than directly aggregating into a ``coarse" class token, 
we introduce the Attribute-Aware Proxy (AAP) embedding module to directly extract attribute features from a single-frame or multi-frame patch token sequence.
Practically, AAP embeddings are formed by a learnable matrix with anisotropic initialization for the richness of learned attributes. 
It can be considered as the ``attribute bank" to serve as the query of the attention operation to match with the feature representations of the input patch tokens.
Specifically, AAP embeddings interact with the keys of the patch token sequence. Finally, the resulting attention map is used to re-weight the value, generating the attribute representations of the specific video clip with the same dimension of AAPs. 
Formally, an AAP embedding module can be written as follows,
\begin{gather}
\textbf{Q}, \ \textbf{K}, \ \textbf{V} = \textbf{P}_Q, \ \textbf{S}\textbf{W}_k,\ \textbf{S}\textbf{W}_v, \notag \\
{\rm AAP}(\textbf{S}) = \frac{{\rm Softmax}(\textbf{QK}^{\rm T})}{\sqrt{d}}\textbf{V}
\end{gather}
here we use the multi-head version of AAP embedding module in practice, which has the same multi-head setting as ${\rm SA(\cdot)}$ in Eqn.(\ref{eq:1}). 
Note that the spatio-temporal input $\textbf{S}$ here can also be $\hat{\textbf{S}} \in \mathbb{R} ^ {\hat{N} \times d}$ for spatial-only use.
Compared with ${\rm SA(\cdot)}$, the newly proposed AAP module consider the query $\textbf{Q}$ in Eqn.(\ref{eq:1}) as the a set of learnable parameters $\textbf{P}_Q \in \mathbb{R} ^ {N_a \times d^{\prime}}$, 
where $N_a \ll N$ is a hyper-parameter that indicates the number of AAPs. 
By controlling $N_a$, the AAP module could have a manually defined capacity, which leads to flexibility for various real applications. 

%

As shown in Fig.~\ref{fig:frmework}, both \textbf{Stage I} and \textbf{Stage III} employ the proposed AAP embedding modules. 
Specifically, in \textbf{Stage I}, the proposed AAP module is firstly used to generate attribute representations from a multi-frame sequence of patch tokens $\textbf{S} \in \mathbb{R} ^ {\hat{T} \times \hat{N} \times d}$ for similarity measurement. 
Although we do not have any attribute-level annotations, we hope the AAP module can automatically learn a rich set of implicit attributes from the entire training dataset,
while these resultant attribute representations could also present discriminative power complementary to ID-only representations.
%
%
%
%
%
To achieve this goal, the ID-level supervision signal is first imposed on the combination of learned attribute representations to constrain its discriminative power. 
In addition, we initialize the AAPs with anisotropic distributions to capture diverse implicit attribute representations. 
In practice, we surprisingly find that such anisotropy can maintain after the model training, which means such optimized AAP could respond to a  set of differentiated attributes. 
%
%
Moreover, the number of AAPs can be relatively large compared with the class token to cover rich attribute information. 
In this sense, both the richness and diversity of learned implicit attributes can be guaranteed. 

In \textbf{Stage III}, we further insert two intra-frame AAP embedding modules before and after the temporal attention of each STA to conduct attribute-aware temporal interaction.  
%
Such a modified STA block is named A-STA, which is illustrated in Fig.~\ref{fig:sta}. 
In A-STA, semantic-related attributes in different frames experience inter-frame interaction to model their temporal relations. In the end, after temporal attention, we set $N_a$ equal to $N$ for the second AAP embedding module so that it has $N$ tokens as output to keep the input-output consistency. 
%
%
\subsection{Identity-Aware Proxy Embedding Module} \label{sec:iap}
Extracting discriminative identity representation is also crucial for video-based re-ID.
%
%
To this end, the Identity-Aware Proxy (IAP) embedding module is proposed for effective and efficient discriminative representation generation. 
In previous works, joint space-time attention has shown promising results~\cite{arnab2021vivit, bertasius2021space}, as it accelerates information aggregation by applying self-attention over spatial and temporal dimensions jointly. 
However, the quadratic computational overheads limit its applicability. 
The IAP embedding module is proposed to address such an issue, which performs joint space-time attention with high efficiency while maintaining the discrimination of the identity feature representation. 

\begin{figure}[t]
	\begin{center}
		\includegraphics[width=0.6\linewidth]{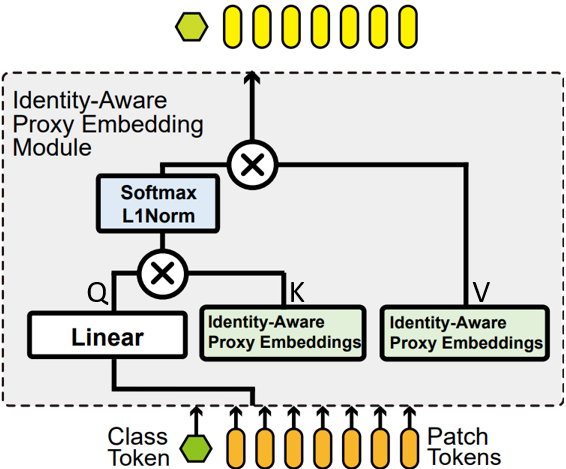}
	\end{center}
	\caption{The detailed module design of the Identity-Aware Proxy (IAP) embedding module. The IAP
    embedding denotes the learnable matrix used to calculate the key or value of the attention operation.
    Here we only show the single-head version of the IAP embedding module and omit the scaling operation. 
    In such a scheme, The output token sequence can be considered as reconstruction by a group of IAPs, which tend to reserve the most discriminative identity features. } 
\label{fig:identity-aware}
\end{figure}

The IAP module contains a set of identity prototypes, which are presented as two learnable matrics. 
In practice, we exploit them to replace the keys $\{p_K^{i}\}_{i=1}^M \in \textbf{P}_K$ and values $\{p_V^{i}\}_{i=1}^M \in \textbf{P}_V$ of the attention operation.
Both $\textbf{P}_K, \textbf{P}_V \in \mathbb{R} ^ {M \times d^{\prime}}$, where $M \in \mathbb{N}^+$ denotes the number of identity prototypes and determine the capacity of the IAP module (usually $M \ll N$). 
As shown in Fig.~\ref{fig:identity-aware}, an attention map $\textbf{A} \in \mathbb{R} ^ {M \times N}$ is first calculated to present the affinity between prototype-patch pairs.
Thus each element in $\textbf{A}$ reflects how close a patch token is to a specific identity prototype.
%
%
%
%
Then this attention map is sparsified by successively applying an L1 normalization and softmax normalization along $M$ and $N$, respectively. 
%
%
%
At last, the class token $\textbf{c}$, \textit{i.e.} the first row of $\textbf{V}$, is updated by the multiplication of $\textbf{V}$ and $\textbf{A}$.
Such an operation aggregates the most discriminative identity features from the entire patch token sequence. 
Formally, given the spatio-temporal token sequence $\textbf{S}$, the output of the IAP module can be calculated as follows:
\begin{gather}
\textbf{Q}, \ \textbf{K}, \ \textbf{V} = \textbf{S}\textbf{W}_q, \textbf{P}_K, \textbf{P}_V                  \notag \\
\textbf{A} = \frac{{\rm Softmax}({\rm L1Norm}(\textbf{QK}^{\rm T}))}{\sqrt{d}} \notag \\
{\rm IAP}(\textbf{S}) = \textbf{AV} \label{eq:5}
\end{gather}
where $\textbf{K}$ and $\textbf{V}$ are not conditioned on input $\textbf{S}$ but are learnable parameters. 
Here we insert an L1 normalization layer before the softmax operation in Eqn. (\ref{eq:5}), resulting in double normalization~\cite{guo2021pct, guo2021beyond}. 
Such a scheme performs patch token re-coding to reduce the noise of patch representations, leading to robust identification results.
Specifically, the learnable matrix $\textbf{P}_K$ matches the input tokens through the double normalization operation to generate the affinity map $\textbf{A}$.
Then these input tokens are thereupon re-coded through a projection of $\textbf{P}_V$ along $\textbf{A}$. 
Since the numbers of learnable vectors in $\textbf{P}_K$ and $\textbf{P}_V$ are much smaller than the number of input tokens, 
the above operation has been able to represent each token in a more compact space (\textit{i.e.} linear combination of the vectors in $\textbf{P}_V$),  effectively suppressing irrelevant information for re-ID.   
%
%
%
%
%
%
%
Moreover, ${\rm IAP}(\cdot)$ has $O(N)$ computational complexity since the number of identity prototypes $M$ is fixed and is usually much less than the total number of patch tokens of a specific video tracklet (\textit{e.g.,} $64$ in our experiments).
So, the proposed IAP embedding module allows all spatio-temporal patch tokens to be processed in parallel for effective and efficient feature extraction. 


\subsection{Temporal Patch Shuffling}\label{sec:ts}
\begin{figure}[t]
	\begin{center}
		\includegraphics[width=1.0\linewidth]{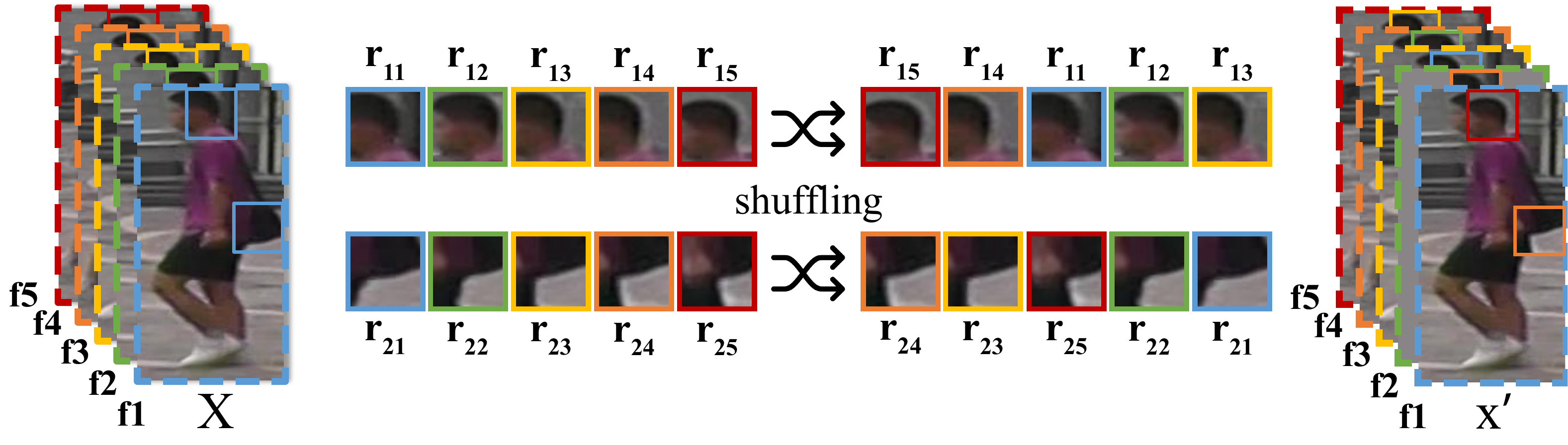}
	\end{center}
	\caption{Visualization of Temporal Patch Shuffling (TPS). $\mathbf{f_{t}}$ represents $t^{th}$ frame, $\mathbf{r}_{\rm it}$ the patch in spatial position $i$ and $t^{th}$ frame. TPS is a built-in data augmentation scheme that randomizes the order of a patch sequence sampled from spatial position $i$. As a result, for example, the patch in the red box is transferred from the $5^{th}$ frame to the $1^{st}$ frame.
	} 
	\label{fig:shuffle}
\end{figure}

To improve the robustness of the model, we propose a novel data augmentation scheme termed Temporal Patch Shuffling (TPS). 
Suppose we have one patch sequences $\mathbf{R}_i=\{\mathbf{r}_{i1},...,\mathbf{r}_{it},...,\mathbf{r}_{iT}\}$ from the same video clip, where the sub-index $i$ denotes specific spatial locations.
As shown in Fig.~\ref{fig:shuffle}, the proposed TPS randomly permutes the patch tokens in $\mathbf{R}_i$ and refill the shuffled sequence $\hat{\mathbf{R}_i}$ to form the new video clip for training.
As illustrated in Fig.~\ref{fig:shuffle}, we could simultaneously select multiple spatial regions in one video clip for shuffling. 
While in the inference phase, the original video clip is directly fed into the model for identification.
TPS brings firm appearance shifts and motion changes from which the network learns to extract generalizable and invariant visual clues. 
In addition, the scale of available training data can be greatly extended based on such a scheme, which helps to prevent the network from overfitting. 

In our experiments, we treat TPS as a plug-and-play operation and implement it at the stem of the network to promote the entire network for the best performance. 
The following section will conduct ablation studies to explore where to insert TPS and to what extent TPS should be for optimal training results.

%

%

%

%

%

\begin{table*}[t]
\centering
\footnotesize
\title{}{}
\begin{tabular}{ccc|ccc|ccc|cc} \\
\toprule[1.5pt]
\multirow{2}{*}{\multirow{1}{*}{Method}} & \multirow{2}{*}{Source}  & \multirow{2}{*}{Backbone}  & \multicolumn{3}{c}{MARS}     & \multicolumn{3}{c}{Duke-V}  & \multicolumn{2}{c}{iLIDS-VID}\\\cline{4-11} 
                                       &              &                & Rank-1 & Rank-5   & mAP                  & Rank-1  & Rank-5 & mAP              & Rank-1        & Rank-5        \\  \hline      
SCAN~\cite{zhang2019scan}              & TIP19      & Pure-CNN        & 87.2 & 95.2 & 77.2                       & -       & -      & -                & 88.0          & 96.7          \\
VRSTC~\cite{hou2019vrstc}              & CVPR19     & Pure-CNN        & -    & 89.8 & 85.1                       & 96.9    & -      & 96.2             & 86.6          & -             \\
M3D~\cite{hou2019vrstc}                & AAAI19     & Pure-CNN        & -    & -    & -                          & 96.9    & -      & 96.2             & 74.0          &94.3             \\
MG-RAFA~\cite{zhang2020multi}          & CVPR20     & Pure-CNN        & 88.8 & 97.0 &\underline{85.9}            & -       & -      & -                & 88.6          & 98.0          \\
AFA~\cite{chen2020temporal}            & ECCV20     & Pure-CNN        & 90.2 & 96.6 & 82.9                       & -       & -      & -                & 88.5          & 96.8          \\
AP3D~\cite{C:gu2020appearance}         & ECCV20     & Pure-CNN        & 90.7    & - & 85.6                       & 97.2    & -      & 96.1             & 88.7          & -             \\
TCLNet~\cite{chen2020temporal}         & ECCV20     & Pure-CNN        & 89.8 & -    & 85.1                       & 96.9    & -      & 96.2             & 86.6          & -             \\

A3D~\cite{J:chen2020learning}          & TIP20     & Pure-CNN        & 86.3  & 95.5       & 80.4          & -    & -     & -     & 86.7  & \underline{98.6}  \\

GRL~\cite{liu2021watching}             & CVPR21    & Pure-CNN        & 90.4 & 96.7 & 84.8                       & 95.0    & 98.7   & 93.8             & 90.4          & 98.3 \\ 
STRF~\cite{aich2021spatio}             & ICCV21    & Pure-CNN        & 90.3  & -       & 86.1          & \underline{97.4}    & -     & 96.4     & 89.3      & -  \\

Fang \MakeLowercase{\textit{et al.}~\cite{fang2021set}} & WACV21 & Pure-CNN & 87.9 & 97.2 & 83.2 &- &- &- & 88.6 & \underline{98.6} \\

TMT~\cite{liu2021video}                & Arxiv21   & CNN-Transformer & 91.2  & \underline{97.3} & 85.8          & -       & -      & -                & \underline{91.3}          & \underline{98.6} \\

Liu \MakeLowercase{\textit{et al.}}~\cite{liu2021video2} & CVPR21*  &CNN-Transformer &\underline{91.3} &- &\textbf{86.5} &96.7 &- &96.2 &- &- \\

STT~\cite{zhang2021spatiotemporal}     & Arxiv21   & CNN-Transformer & 88.7 & -    & \underline{86.3}              & \textbf{97.6} & - & \textbf{97.4}  & 87.5          & 95.0 \\ 

ASANet~\cite{chai2022video}            & TCSVT22   & Pure-CNN        & 91.1 & 97.0 & 86.0                       & \textbf{97.6}    & \textbf{99.9}   & \underline{97.1}           & -             & -             \\ \hline

MSTAT(ours)                            & -         & Pure-Transformer & \textbf{91.8} & \textbf{97.4} & 85.3  & \underline{97.4} & \underline{99.3} & 96.4              & \textbf{93.3} & \textbf{99.3}    \\ 
\bottomrule[1.5pt] 
\end{tabular}
\caption{Result comparison with state-of-the-art video-based person re-ID methods on MARS, DukeMTMC-VideoReID, and iLIDS-VID. * denotes the workshop of the conference. }
\label{tab:my-table}
\end{table*}

\section{Experiment}
\label{sec:exp}

\subsection{Datasets and evaluation protocols} 
In this paper, we evaluate our proposed MSTAT on three widely-used video-based person re-ID benchmarks: iLIDS-VID~\cite{wang2014person}, DukeMTMC-VideoReID (DukeV)~\cite{ristani2016performance}, and MARS~\cite{C:zheng2016mars}.  

\textit{1)} iLIDS-VID~\cite{wang2014person} is comprised of $600$ video tracklets of $300$ persons captured from two cameras. In these video tracklets, frame numbers range from $23$ and $192$. The test set shares $150$ identities with the training set. 

\textit{2)} DukeMTMC-VideoReID~\cite{ristani2016performance} is a large-scale video-based benchmark which contains $4,832$ videos sharing $1,404$ identities. In the following sections, we use the abbreviation “DukeV” for the DukeMTMC-VideoReID dataset. The video sequences in the DukeV dataset are commonly longer than videos in other datasets, which contain $168$ frames on average. 

\textit{3)} MARS~\cite{C:zheng2016mars} is one of the largest video re-ID benchmarks which collects $1,261$ identities existing in around $20, 000$ video tracklets captured by $6$ cameras. Frames within a video tracklet are relatively more misaligned since they are obtained by a DPM detector\cite{felzenszwalb2009object} and a GMMCP tracker~\cite{dehghan2015gmmcp} rather than hand drawing. Furthermore, around $3,200$ distractor tracklets are mixed into the dataset to simulate real-world scenarios.

For evaluation on MARS and DukeV datasets, we use two metrics: the Cumulative Matching Characteristic (CMC) curves~\cite{bolle2005relation} and mean Average Precision (mAP) following previous works~\cite{zhang2019scan, zhang2020multi, chen2020temporal, liu2021video}.
However, in the gallery set of iLIDS-VID, there is merely one correct match for each query. For this benchmark, only cumulative accuracy is reported. 

\subsection{Implementation details} 
Our proposed MSTAT framework is built based on Pytorch toolbox~\cite{paszke2019pytorch}. In our experiments, it is running on a single NVIDIA A$100$ GPU ($40$G memory). We resize each video frame to $224 \times 112$ for the above benchmarks. Typical data augmentation schemes are involved in training, including horizontal flipping, random cropping, and random erasing. For all stages, STA modules are pretrained on an action recognition dataset, K600~\cite{carreira2018short}, while other aforementioned modules are randomly initialized. 
%

In the training phase, if not specified, we sample $L = 8$ frames each time for a video tracklet and set the batch size as 24. In each mini-batch, we randomly sample two video tracklets from different cameras for each person. We supervise the network by cross-entropy loss with label smoothing~\cite{szegedy2016rethinking} associated with widely used BatchHard triplet loss~\cite{hermans2017defense}. Specifically, we impose supervision signals separately on the concatenated attribute representation from the AAP embedding module in \textbf{Stage I}, the output class tokens from \textbf{Stage II}, and \textbf{Stage III}. The learning rate is initially set to 1e-3, which would be multiplied by $0.75$ after every $25$ epochs. The entire network is updated by an SGD optimizer in which the weight decay and Nesterov momentum are set to $5 \times 10^{-5}$ and $0.9$, respectively. 

In the test phase, following~\cite{zhang2021spatiotemporal, C:gu2020appearance}, we randomly sample 32 frames as a sequence from each original tracklet in either query or gallery. For each sequence, The attribute representation from \textbf{Stage I}, the output class tokens from stage \textbf{Stage II} and \textbf{Stage III} are concatenated as the overall representation. Following the widely-used protocol, we compute the cosine similarity between each query-gallery pair using their overall representations. Then, the CMC curves and the mAP can be calculated based on the predicted ranking list and the ground truth identity of each query. Note that we do not use any re-ranking technique. 

%

\subsection{Compared with the state of the arts} 
In Table~\ref{tab:my-table}, we make a comparison on three benchmark datasets between our method and video-based person re-ID methods from 2019 to 2021, including M3D~\cite{hou2019vrstc}, GRL\cite{liu2021watching}, STRF~\cite{aich2021spatio}, Fang \MakeLowercase{\textit{et al.}}~\cite{fang2021set}, TMT~\cite{liu2021video}, Liu \MakeLowercase{\textit{et al.}}~\cite{liu2021video2}, ASANet~\cite{chai2022video}. According to their backbones, these re-ID methods can be roughly divided into the following types: Pure-CNN, CNN-Transformer Hybrid, and Pure-Transformer methods. 
In real-world applications, rank-1 accuracy~\cite{bolle2005relation} reflects what extent a method can find the most confident positive sample~\cite{ye2021deep}, and relatively high rank-1 accuracy can save time in confirmation. As the first method based on Pure-Transformer for video-based re-ID so far, we achieve state-of-the-art results in rank-1 accuracy on three benchmarks. Our approach especially attains rank-1 accuracy of $91.8\%$ and rank-5 accuracy of $97.4\%$ on the largest-scale benchmark, MARS. It is noteworthy that our MSTAT outperforms the best pure CNN-based methods using ID annotations only by a margin of $1.1\%$ and a CNN-Transformer hybrid method, TMT, by $0.6\%$ in MARS rank-1 accuracy. 

Compared to our proposed method, TCLNet~\cite{chen2020temporal} explicitly captures complementary features over different frames, and GRL~\cite{liu2021watching} devises a guiding mechanism for reciprocating feature learning. 
However, the designed modules in these methods commonly take as input the deep spatial feature maps extracted by a CNN backbone (\textit{e.g.} ResNet50) that may overlook attribute-associated or identity-associated information without explicit modeling.
Similar to ours, TMT~\cite{liu2021video} and M3D~\cite{aich2021spatio} process video tracklets in multiple views to extract and fuse multi-view features. 
Notably, in all stages of MSTAT, intermediate features are spatio-temporal and can be iteratively updated to capture spatio-temporal cues with different emphases.  %
ASANet~\cite{chai2022video} exploits explicit ID-relevant attributes (\textit{e.g.}, \textit{gender, clothes, and hair}) and ID-irrelevant attributes (\textit{e.g.}, \textit{pose and motion}) on a multi-branch network. 
Despite the performance growth, the demand for attribute annotations may limit its applications in large-scale scenarios. 
In comparison with existing methods, our method aggregates spatio-temporal information in a unified manner and explicitly capitalizes on implicit attribute information to improve recognizability under challenging scenarios. 
Conclusively, our method achieves the state-of-the-art performance of $91.8\%$ and $93.3\%$ rank-1 accuracy, respectively, on MARS and iLIDS-VID.
%


\subsection{Effectiveness of Multi-Stage Framework Architecture}
To evaluate the effectiveness of the three stages in our proposed MSTAT, we carry out a series of ablation experiments whose results are displayed in Table~\ref{tab:stages}. After the three stages are jointly trained, we first separately evaluate each stage using its output feature representation. Then, we concatenate two or more stages to evaluate whether each is effective. 
%

For three single stages, each has rank-1 accuracy ranging from $89.2$ to $89.8$. However, their combinations result in a significant increase of over $0.8\%$. Remarkably, while \textbf{Stage I} and \textbf{Stage II} secure only $89.2$ rank-1 accuracy, their integration attains up to $91.2\%$, surpassing them by a $2\%$ margin. One can attribute such a result to their emphases: one stage on attribute-associated features and the other on identity-associated features. Eventually, when all three stages are used, MSTAT reaches a $91.8\%$ rank-1 accuracy, higher than all two-stage combinations. Overall, these experiments demonstrate that the three stages have different preferences toward features and can complement each other by simple concatenation.

\subsection{Effectiveness of Key Components}
To demonstrate the effectiveness of our proposed MSTAT, we conduct a range of ablative experiments on the largest public benchmark MARS. 

\subsubsection{Effectiveness of Attribute-Aware Proxy Embedding Module} \label{exp:1}
%

\begin{table}\small\centering

	\begin{tabular}{c|c|ccc}
		\toprule[1.5pt]
		
		\multirow{1}*{Method}  &Test Protocol        &Rank-1  &Rank-5  &mAP\\
		\hline
		\multirow{7}*{MSTAT}   &Stage \uppercase\expandafter{\romannumeral1}              &89.2    &96.7    &82.4\\
		\multirow{0}*{}        &Stage \uppercase\expandafter{\romannumeral2}               &89.2    &96.5    &83.0\\
		\multirow{0}*{}        &Stage \uppercase\expandafter{\romannumeral3}               &89.8\   &96.5    &83.0\\
		\multirow{0}*{}        &Stage \uppercase\expandafter{\romannumeral1} \& \uppercase\expandafter{\romannumeral2}            &91.2    &97.3    &85.0\\
		\multirow{0}*{}        &Stage \uppercase\expandafter{\romannumeral1} \& \uppercase\expandafter{\romannumeral3}              &90.5    &97.2    &83.9\\
		\multirow{0}*{}        &Stage \uppercase\expandafter{\romannumeral2} \& \uppercase\expandafter{\romannumeral3}              &90.6    &96.9    &84.6\\
		\multirow{0}*{}        &Stage \uppercase\expandafter{\romannumeral1}, \uppercase\expandafter{\romannumeral2}, \& \uppercase\expandafter{\romannumeral3}          &\textbf{91.8} &\textbf{97.4} &\textbf{85.3}\\
		
		\bottomrule[1.5pt]
\end{tabular}
\caption{Ablation study on three stages of MSTAT on MARS. Test Protocol means the final feature representation used for similarity measurement. The network architecture and training hyper-parameter setting remain the same for each experiment.} 
\label{tab:stages}
\end{table}
%

		
%

\begin{figure}[t]
	\begin{center}
		\includegraphics[width=\linewidth]{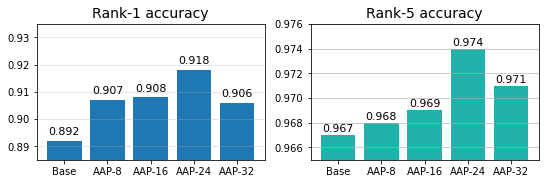}
	\end{center}
	\caption{Ablation study on the attribute-aware proxy (AAP) embedding module for attribute extraction in MARS. "Base" is the network without attribute extraction using AAP in training and testing. AAP-$\mathbf{k}$ indicates the network where the AAP embedding module in \textbf{Stage I} has $\mathbf{k}$ AAPs.} 
	\label{fig:apm_abl}
\end{figure}

%
As shown in Fig.~\ref{fig:apm_abl}, we evaluate MSTAT with different AAP numbers (\textit{i.e.} $N_a$ in Sec.~\ref{sec:aap}) in the AAP embedding module in the last layer of \textbf{Stage I}. The figure reveals that 24 proxies are optimal for attributive information extraction as it attains the best performance in terms of rank-1 and rank-5 accuracy. In contrast to the baseline, MSTAT has seen over $2\%$ growth in rank-1 accuracy and around $1\%$ in rank-5 accuracy. However, a redundant or insufficient number of AAPs may cause a minor performance drop since they may pay attention to noisy or useless attributes. In summary, the AAP embedding module for clue extraction gives a boost to the performance in rank-1 and rank-5 accuracy, with negligible computational overhead.

\begin{figure}[t]
	\begin{center}
		\includegraphics[width=\linewidth]{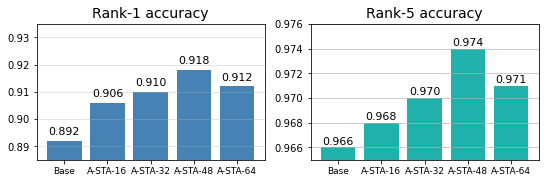}
	\end{center}
	\caption{Ablation study on A-STA. "Base" is the network that consists of STA only. A-STA-$\mathbf{k}$ represents the network in which \textbf{Stage III} is equipped with A-STA layers each of $\mathbf{k}$ AAPs.} 
	\label{fig:a-sta_abl}
\end{figure}

Attribute-Aware Proxy (AAP) embedding modules are also used for A-STA, a variant of STA for attribute-aware temporal feature fusion in \textbf{Stage III}. As shown in Fig.~\ref{fig:a-sta_abl}, we conduct a series of experiments to explore whether A-STA is effective and how many AAPs for A-STA are appropriate (also corresponding to $N_a$ in \ref{sec:aap})). The experiment results reveal that the baseline model fails to reach $90\%$ rank-1 accuracy or $97\%$ rank-5 accuracy. As the number of AAPs increases, these two metrics grow to 91.8\% and 97.4\%. 
\begin{figure}[t]
	\begin{center}
		\includegraphics[width=\linewidth]{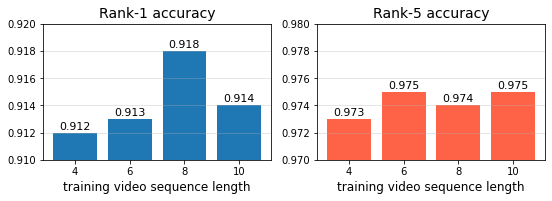}
	\end{center}
	\caption{Study on the effect of training video sequence length on MARS.} 
	\label{fig:seq_len}
\end{figure}

Therefore, we can attribute the performance soar to A-STA, allowing for attribute-aware temporal interaction. A-STA offers a different viewpoint from that of \textbf{Stage II} on videos. Moreover, due to the redundancy of temporal information in many video re-ID scenarios discussed in~\cite{chen2020temporal}, A-STA with too many AAPs incurs meaningless attributes. This can be why the performance descends once A-STA has too many AAPs. 
%
%

In conclusion, our proposed AAP embedding module can be used for: (1) the extraction of informative attributes as plugged into any Transformer layer and (2) attribute-aware temporal interaction when a temporal attention module is sandwiched between two. Both of the two functionalities cause a significant increase in performance, demonstrating their effectiveness.

\subsubsection{Effectiveness of Identity-Aware Proxy Embedding Module}

%


In Table~\ref{tab:iap_pos}, MSTAT that discards IAP embedding modules leads to only $88.2\%$ rank-1 accuracy and $96.4\%$ rank-5 accuracy. However, it boosts rank-1 performance by $2.8\%$ or $2.2\%$ by taking the place of STA in \textbf{Stage II} or \textbf{Stage III}. Finally, IAP embedding modules in the last layers in both \textbf{Stage II} and \textbf{Stage III} further improve $0.8\%$ rank-1 accuracy and $0.4\%$ rank-5 accuracy. The IAP embedding module's ablation results demonstrate its ability to generate discriminative representations efficiently. Intuitively, we place the IAP embedding module only in the last few depths because it may discard non-discriminative features that should be preserved in shallow layers. 
%


%

%


\subsubsection{Effectiveness of Temporal Patch Shuffling}
To evaluate the effectiveness of Temporal Patch Shuffling (TPS), we assign different probabilities to implement TPS for each training video sample. Note that in the following experiments, the number of spatial positions to shuffle is set to $5$ if we implement TPS on this sample. As shown in Table~\ref{tab:ps}, $20\%$ probability provides the best result over others, which leads to a growth of $0.3\%$ in rank-1 accuracy. However, the $60\%$ or $80\%$ probability results in a $0.1\%$ or $0.2\%$ rank-1 accuracy drop mainly due to heavy noise. In summary, a proper level of TPS would be an effective data augmentation method for the Transformer for video-based person re-ID. Further, rather than reserving temporal motion (an ordered sequence of patches), TPS stimulates re-identification accuracy by learning temporal coherence from shuffled patch tokens. 


		

\begin{table}\scriptsize\centering 

	\begin{tabular}{c|c|ccc}
		\toprule[1.5pt]
		
		\multirow{1}*{Method}                     &Position      &Rank-1   &Rank-5\\
		\hline
		\multirow{1}*{w/o IAP embedding module}   &-             &88.2     &96.4\\
		\hline
		\multirow{3}*{w/ IAP embedding module}    &Stage II      &91.0     &97.0\\
		                                           &Stage III     &90.4     &97.0\\
		                                           &Stage II\&III &\textbf{91.8} &\textbf{97.4}\\
		\bottomrule[1.5pt]
	\end{tabular}
	\caption{Ablation study on the IAP embedding module. Stage \uppercase\expandafter{\romannumeral2} and Stage \uppercase\expandafter{\romannumeral3} in this table means that an IAP embedding module is appended to the last layer of Stage \uppercase\expandafter{\romannumeral2} and Stage \uppercase\expandafter{\romannumeral3} respectively. This table shows that the IAP embedding module brings improvements to every single stage. When it is placed on both two stages, MSTAT shows the best performance.} 
\label{tab:iap_pos}
\end{table}
\begin{table}\small\centering
	\begin{tabular}{c|c|ccc}
		\toprule[1.5pt]
		
		\multirow{1}*{Methods}&Prob.  &Rank-1   &Rank-5   &mAP\\
		\hline
		\multirow{1}*{MSTAT w/o TPS} &0\%    &91.5     &\textbf{97.5}     &85.2\\
		\hline
		\multirow{4}*{MSTAT w/ TPS } &20\%  &\textbf{91.8}  &97.4  &\textbf{85.3}\\
		                      &40\%  &91.7     &97.3     &85.2\\
		                      &60\%  &91.4     &\textbf{97.5}     &85.1\\
		                      &80\%  &91.3     &97.1     &85.1\\
		\bottomrule[1.5pt]
\end{tabular}
\caption{Ablation study on Temporal Patch Shuffling. The table shows that the proper level of shuffling can bring slight improvement. However, it may degrade the learning while the shuffling degree becomes increasingly overwhelming.} 
\label{tab:ps}
\end{table}

\subsection{Effect of video sequence length}
To investigate how temporal noise influences the training of MSTAT, we conduct experiments on videos with varied lengths. In Fig.~\ref{fig:seq_len}, experiments provide length-varying video tracklets for training, while all experiments are implemented under the identical evaluation setting with a fixed video length of $32$. All experiments shut down until the loss stops decreasing for ten epochs. 

On the one hand, rank-1 accuracy shows an upward trend as temporal noise gradually decreases, reaching a peak at $8$. On the other hand, temporal noise shows no apparent correlation with rank-5 accuracy and mAP. These results show that our model gains up to $0.6\%$ rank-1 accuracy through learning better temporal features from data. However, rank-5 accuracy and mAP benefit little from noise reduction, from which we can speculate that in most cases in video re-ID, learning temporal features is less important than learning appearance features as they only account for 0.6\% of rank-1 and 0.2\% of rank-5 accuracy. Similar results can be found in ~\cite{liu2021video}.

\subsection{Comparison among metric learning methods}
Metric learning aims to regularize the sample distribution on feature space. Usually, metric learning losses constrain the compactness of intra-class distribution and sparsity of the overall distribution. To explore which strategy cooperates with our framework better, we compare a range of classic metric learning loss functions on iLIDS-VID, as shown in Table \ref{tab:metric}. Note that these losses are scaled to the same magnitude to ensure fairness. Significantly, OIM loss~\cite{xiao2017joint} and BatchHard triplet loss~\cite{hermans2017defense}, widely used in re-ID, outperform Arcface~\cite{deng2019arcface} and SphereFace~\cite{liu2017sphereface} losses by a large margin since the latter two loss functions suffer from untimely overfitting in our experiments.

\begin{table}\footnotesize\centering
	\begin{tabular}{c|cc}
		\toprule[1.5pt]
		
		\multirow{1}*{Metric learning loss}  &Rank-1   &Rank-5\\
		\hline
		\multirow{1}*{w/o Metric learning}   &66.0     &90.0 \\
		\multirow{1}*{Arcface~\cite{deng2019arcface}}       &73.3     &90.7\\
		\multirow{1}*{SphereFace~\cite{liu2017sphereface}}  &66.7     &89.3\\
		\multirow{1}*{OIM~\cite{xiao2017joint}}             &89.3     &98.3\\
		\multirow{1}*{BatchHard* ~\cite{hermans2017defense}}  & \textbf{93.3}  & \textbf{99.3}\\
		\bottomrule[1.5pt]
\end{tabular}
\caption{
Comparison among metric learning loss functions on iLIDS-VID, where * denotes the method used in our implementation. For Arcface and SphereFace, we test three margins and report the best result: (1) by default, (2) 20\% larger than the default, (3) 20\% smaller than the default.  
}
\label{tab:metric}
\end{table}

        %

    			


		

\subsection{Visualization}


\begin{figure}[t]
	\begin{center}
		\includegraphics[width=0.8\linewidth]{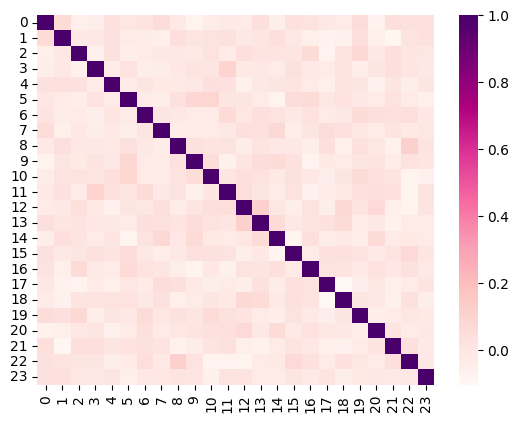}
	\end{center}
	\caption{Visualization of the similarity matrix of attribute-aware proxies trained on MARS. The maximal similarity between all pairs is around 0.2, demonstrating that AAPs learn to capture diverse attributes. 
	} 
	\label{fig:cosine_mat}
\end{figure}

%
To better understand how the proposed framework works, we conduct visualization on the AAP embedding module. In Fig.~\ref{fig:cosine_mat}, we show the diversity of implicit attributes by the similarity matrix of 24 AAPs. This figure implies that AAPs are anisotropic, covering different attribute features that appear in the given training dataset. 

\begin{figure}[t]
	\begin{center}
		\includegraphics[width=1.0\linewidth]{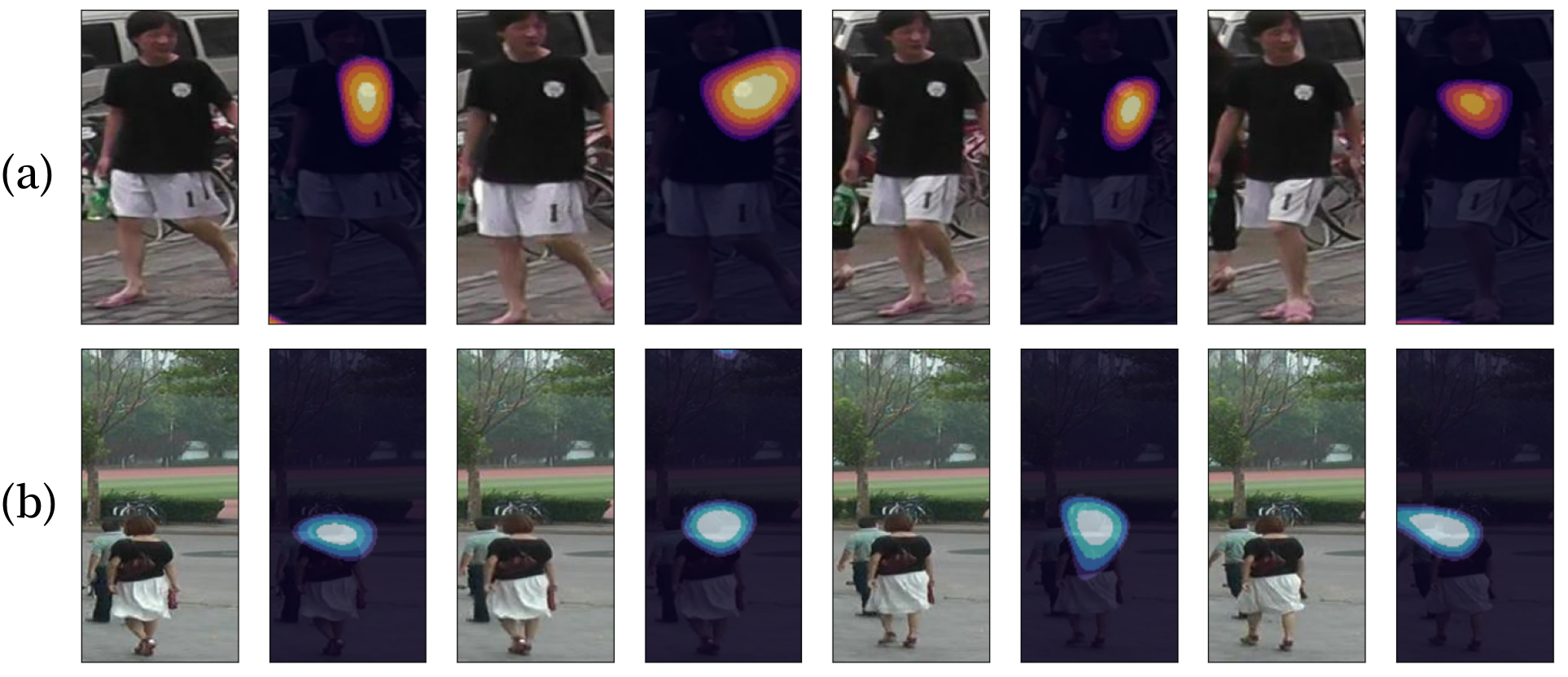}
	\end{center}
	\caption{Visualization of attribute-aware proxies for two different pedestrians on MARS. Attention heat maps of four consecutive frames from the AAP embedding module on Stage I are displayed. 
	} 
	\label{fig:vis}
\end{figure}

Specifically, as shown in Fig.~\ref{fig:vis}, we randomly select two pedestrians' tracklets. Attention map visualization is adopted as a sign of each AAP's concentration. In practice, we process the raw attention maps first by several average filters and then by thresholding to deliver smooth visual effects instead of grid-like maps. In these heap maps, the brighter color denotes the higher attention value. 
Despite the absence of attribute-level supervision, Fig.\ref{fig:vis} shows that some AAPs learn to pay attention to a local region with special meanings as an identity cue. For example, the AAP in white color in video clips (a) automatically learns to cover the logo in the T-shirt, while the one in (b) captures the head of the woman.

\begin{figure}[t]
	\begin{center}
		\includegraphics[width=1.01\linewidth]{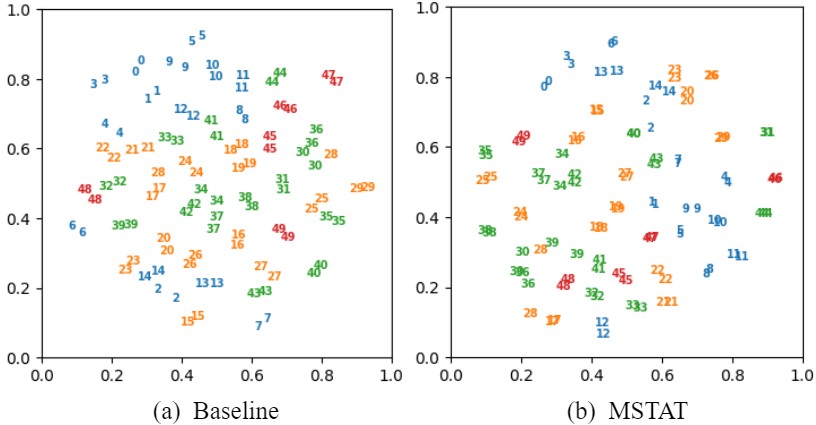}
	\end{center}
	\caption{T-SNE Visualization of the iLIDS-VID test set. The numbers on the plots indicate person IDs. MSTAT shows an increase in intra-class compactness and the minimum inter-class distance over the entire test set compared to the baseline. 
	} 
	\label{fig:t-sne}
\end{figure}

Moreover, we display the t-SNE visualization result on iLIDS-VID in Fig. \ref{fig:t-sne}. 
It only contains the first 1/3 of the IDs in the test set for a better visual effect. 
We also provide the corresponding quantitative evaluation results in Table~\ref{tab:ilids} measured by the normalized averaged intra-class distance and the minimum inter-class distance (0-2) on the entire test set. 
As a result, MSTAT drops the average intra-class distance from 0.4572 of the baseline to 0.4410 and enlarges the minimum inter-class distance from 0.4704 to 0.5012. 
Further, to eliminate the influence of accuracy, we measure the intra-class distance between correctly matched samples, from which we witness a similar result. 
These results explain why MSTAT's t-SNE visualization seems sparser.

\begin{table}\footnotesize\centering
	\begin{tabular}{c|cccc}
		\toprule[1.5pt]
		
		\multirow{1}*{Methods}  & Intra$\downarrow$ & Intra$\downarrow$ & Inter$\uparrow$ & Rank-1$\uparrow$\\
		\hline
		\multirow{1}*{Baseline~\cite{bertasius2021space}}   &0.4572	&0.4495 &0.4704 &0.873 \\
		\multirow{1}*{MSTAT w/o attr.}                      &0.4517 &0.4469 &0.4644 &0.913\\
		\multirow{1}*{MSTAT (ours)}  & \textbf{0.4410} & \textbf{0.4389} & \textbf{0.5012} & \textbf{0.933} \\
		\bottomrule[1.5pt]
\end{tabular}
\caption{Quantitative evaluation on iLIDS-VID. "Intra" denotes the averaged normalized intra-class distance, and "Inter" is the minimum inter-class distance. Here, * means that the metric is computed on samples with the correct rank-1 match. 
}
\label{tab:ilids}
\end{table}

\section{Conclusion}
\label{sec:conclusion}
This paper proposes a novel framework for video-based person re-ID, referred to as Spatial-Temporal Aggregation Transformer (MSTAT). To tackle simultaneous extraction for local attributes and global identity information, MSTAT adopts a multi-stage architecture to extract (1) attribute-associated, (2) the identity-associated, and (3) the attribute-identity-associated information from video clips, with all layers inherited from the vanilla Transformer. Further, for reserving informative attribute features and aggregating discriminative identity features, we introduce two proxy embedding modules (Attribute-Aware Proxy embedding module and Identity-Aware Proxy embedding module) into different stages. In addition, a patch-based data augmentation scheme, Temporal Patch Shuffling, is proposed to force the network to learn invariance to appearance shifts while enriching training data. Massive experiments show that MSTAT can extract attribute-aware features consistent across frames while reserving discriminative global identity information on different stages to attain high performance. Finally, MSTAT outperforms most existing state-of-the-arts on three public video-based re-ID benchmarks.

Future work may focus on mining the hard instances or local informative attribute locations to conduct contrastive learning to promote the model's accuracy further.
Moreover, leveraging more unlabeled and multi-modal data to improve the model's effectiveness is also a potential research direction.



\section*{Acknowledgment}
The work is supported in part by the Young Scientists
Fund of the National Natural Science Foundation of China under grant No. 62106154, by National Key R\&D Program of China under Grant No. 2021ZD0111600, by Natural Science Foundation of Guangdong Province, China (General Program) under grant No.2022A1515011524, by Guangdong Basic and Applied Basic Research Foundation under Grant No. 2017A030312006, by CCF-Tencent Open Fund, by Shenzhen Science and Technology Program ZDSYS20211021111415025, and by the Guangdong Provincial Key Laboratory of Big Data Computing, The Chinese Univeristy of Hong Kong
(Shenzhen).







%
{\small
\bibliographystyle{ieee}
\bibliography{reference}
}

%

\begin{IEEEbiography}[{\includegraphics[width=1in,height=1.25in,clip]{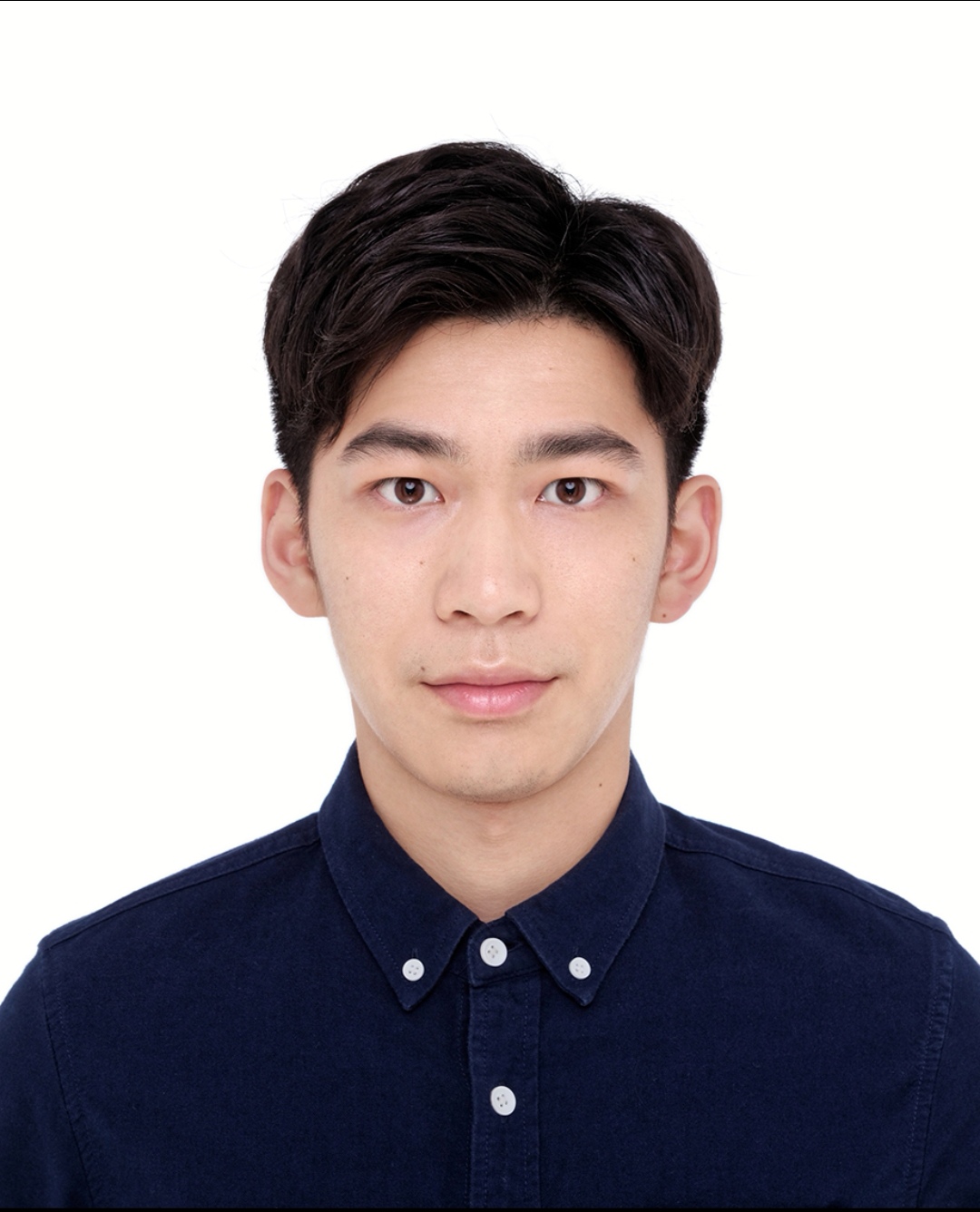}}]{Ziyi Tang} is now pursuing his Ph.D. degree at Sun Yat-Sen University. Before that, he was a research assistant at The Chinese University of Hong Kong, Shenzhen (CUHK-SZ), China. He received the B.E. degree from South China Agriculture University (SCAU), Guangzhou, China in 2019 and M.S. degree from The University of Southampton, Southampton, U.K. in 2020. He has won top places in data science competitions hosted by Kaggle and Huawei respectively. His research interests include Computer Vision, Vision-Language Joint Modeling, and Casual Inference. 
\end{IEEEbiography}

\begin{IEEEbiography}[{\includegraphics[width=1in,height=1.25in,clip]{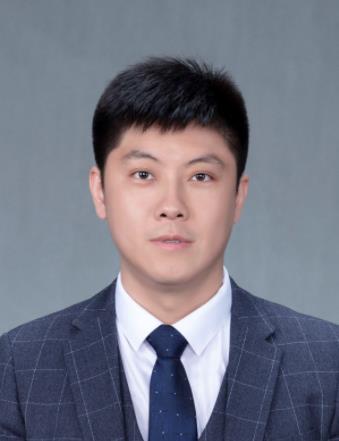}}]{Ruimao Zhang} is currently a Research Assistant Professor in the School of Data Science, The Chinese University of Hong Kong, Shenzhen (CUHK-SZ), China. He is also a Research Scientist at Shenzhen Research Institute of Big Data. He received the B.E. and Ph.D. degrees from Sun Yat-sen University, Guangzhou, China in 2011 and 2016, respectively. 
From 2017 to 2019, he was a Post-doctoral Research Fellow in the Multimedia Lab, The Chinese University of Hong Kong (CUHK), Hong Kong. After that, he joined at SenseTime Research as a Senior Researcher until 2021. 
His research interests include computer vision, deep learning and related multimedia applications. 
He has published about 40 peer-reviewed articles in top-tier conferences and journals such as TPAMI, IJCV, ICML, ICLR, CVPR, and ICCV.
He has won a number of competitions and awards such as Gold medal in 2017 Youtube 8M Video Classification Challenge, the first place in 2020 AIM Challenge on Learned Image Signal Processing Pipeline.
%
%
He was rated as Outstanding Reviewer of NeurIPS in 2021.
He is a member of IEEE.

\end{IEEEbiography}

\begin{IEEEbiography}[{\includegraphics[width=1in,height=1.25in,clip,keepaspectratio]{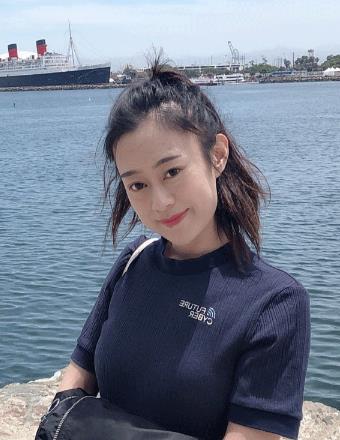}}]{Zhanglin Peng} is now pursuing her Ph.D. degree with the Department of Computer Science, The University
of Hong Kong, Hong Kong, China. She received her B.E. and M.S. degrees from Sun Yat-Sen University, Guangzhou, China in 2013 and 2016, respectively.
From 2016 to 2020, she was a researcher at SenseTime Research.
Her research interests are computer vision and machine learning.
\end{IEEEbiography}

\begin{IEEEbiography}[{\includegraphics[width=1in,height=1.25in,clip,keepaspectratio]{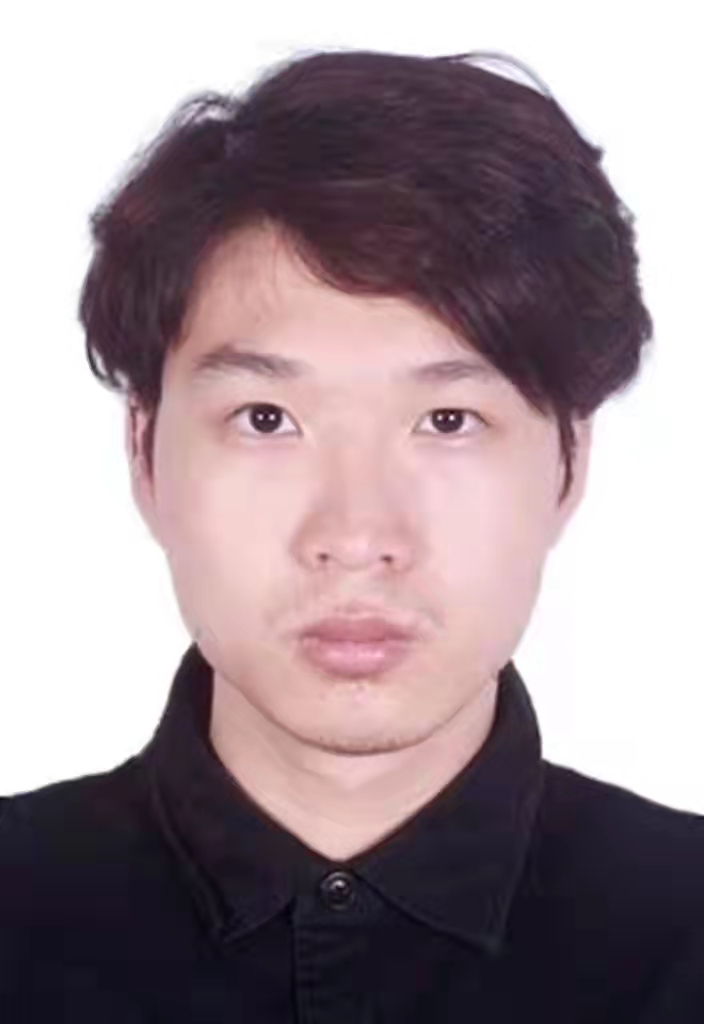}}]{Jinrui Chen} 
is currently pursuing the B.A. degree in Financial Engineering conferred jointly by the School of Data Science, the School of Science and Engineering, and the School of Management and Economics, The Chinese University of Hong Kong, Shenzhen (CUHK-SZ), China. His research interests include deep learning and financial technology.
\end{IEEEbiography}


\begin{IEEEbiography}[{\includegraphics[width=1in,height=1.25in,clip,keepaspectratio]{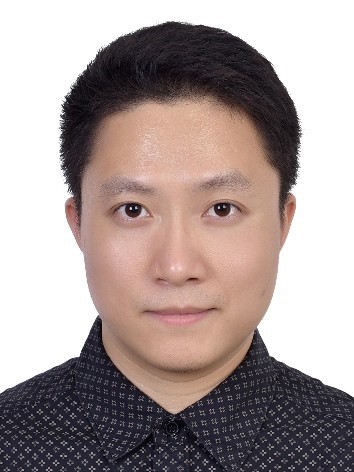}}]{Liang Lin} (M’09, SM’15) is a Full Professor of computer science at Sun Yat-sen University. He served as the Executive Director and Distinguished Scientist of SenseTime Group from 2016 to 2018, leading the R\&D teams for cutting-edge technology transferring. He has authored or co-authored more than 200 papers in leading academic journals and conferences, and his papers have been cited by more than 22,000 times. He is an associate editor of IEEE Trans. Multimedia and IEEE Trans. Neural Networks and Learning Systems, and served as Area Chairs for numerous conferences such as CVPR, ICCV, SIGKDD and AAAI. He is the recipient of numerous awards and honors including Wu Wen-Jun Artificial Intelligence Award, the First Prize of China Society of Image and Graphics, ICCV Best Paper Nomination in 2019, Annual Best Paper Award by Pattern Recognition (Elsevier) in 2018, Best Paper Dimond Award in IEEE ICME 2017, Google Faculty Award in 2012. His supervised PhD students received ACM China Doctoral Dissertation Award, CCF Best Doctoral Dissertation and CAAI Best Doctoral Dissertation. He is a Fellow of IET and IAPR.

\end{IEEEbiography}

\end{document}